\documentclass{article}
\usepackage{spconf,amsmath,graphicx}
\usepackage{amsfonts}
\usepackage{bbding}
\usepackage{booktabs}
\usepackage{tabularx}
\usepackage{textcomp}
\usepackage{xcolor}
\usepackage{booktabs} 
\usepackage{threeparttable}
\usepackage{multirow}

\usepackage{enumitem}
\setlist{nosep, leftmargin=14pt}

\usepackage{mwe} 

\title{Deeply Supervised Layer Selective Attention Network: Towards Label-Efficient Learning for Medical Image Classification}
\name{Peng Jiang, Juan Liu\sthanks{Corresponding author.}, Lang Wang, Zhihui Ynag, Hongyu Dong, Jing Feng}
\address{Institute of Artificial Intelligence, School of Computer Science, Wuhan University, Wuhan, China\\\{pelenjiang,liujuan,fantasy,zhy,HongyuDong,gfeng\}@whu.edu.cn}
\begin{document}
%
\maketitle
\begin{abstract}
Labeling medical images depends on professional knowledge, making it difficult to acquire large amount of annotated medical images with high quality in a short time. 
Thus, making good use of limited labeled samples in a small dataset to build a high-performance model is the key to medical image classification problem. In this paper, we propose a deeply supervised Layer Selective Attention Network (LSANet), which comprehensively uses label information in feature-level and prediction-level supervision. For feature-level supervision, in order to better fuse the low-level features and high-level features, we propose a novel visual attention module, Layer Selective Attention (LSA), to focus on the feature selection of different layers. LSA introduces a weight allocation scheme which can dynamically adjust the weighting factor of each auxiliary branch during the whole training process  to further enhance deeply supervised learning and ensure its generalization. For prediction-level supervision, we adopt the knowledge synergy strategy to promote hierarchical information interactions among all supervision branches via pairwise knowledge matching. Using the public dataset, MedMNIST, which is a large-scale benchmark for biomedical image classification covering diverse medical specialties, we evaluate LSANet on multiple mainstream CNN architectures and various visual attention modules. The experimental results show the substantial improvements of our proposed method over its corresponding counterparts, demonstrating that LSANet can provide a promising solution for label-efficient learning in the field of medical image classification.

\end{abstract}
\section{Introduction}

\begin{figure*}[t]
\centering
\includegraphics[width=0.7\textwidth]{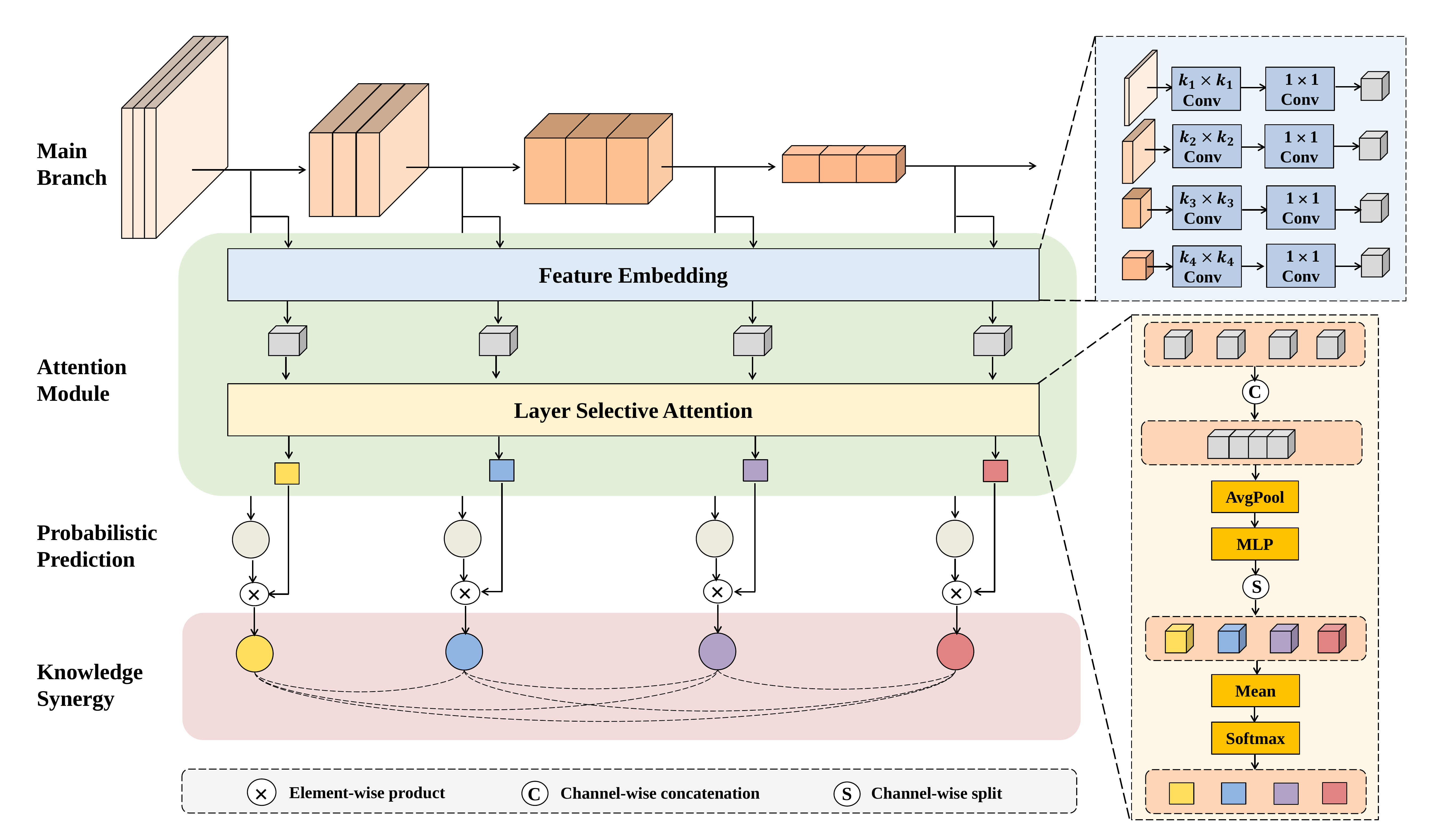} 
\caption{The overview architecture of LSANet.}
\label{fig1}
\end{figure*}

\noindent In recent years, deep learning has achieved remarkable success in computer vision. Deep convolutional neural network (DCNN) based image analysis approaches have consequently blossomed in many medical fields, such as dermatology, pathology and ophthalmology \cite{ref1,ref2,ref3}. 
It is usually requires plenty of well-annotated samples to train a DCNN model with promising performance. However, it is very difficult to obtain enough labeled medical images for two main reasons. First, it is hard to construct a large-scaled dataset with high quality and consistency in a short time due to the rarity of the diseases and the data security issues \cite{ref4}. Second, the annotation of medical images relies heavily on experienced experts with medical expertise. Therefore, how to efficiently utilize the limited labeled samples of a small dataset to build higher-performance models is a crucial problem in the field of DCNN based medical image classification.



It is generally believed that deeper network leads to better performance \cite{refvgg,refgooglenet,refresnet}. Therefore, many researchers try to improve the classification performance by increasing the depth of the network. For example, Yu et al. \cite{ref5} validated that 50-layer FCRN (fully convolutional residual network) outperformed 16-layer VGGNet, 20-layer GoogLeNet and 38-layer FCRN for the Melanoma detection challenge. However, they also found that the model of FCRN-101 performed worse than that of FCRN-50, illustrating that a deeper network structure does not necessarily guarantee better performance in the field of medical image analysis. Similar phenomenon also occurs in some other medical tasks \cite{ref6,ref7,ref8,ref9}. One of the underlying reasons of such phenomenon may be that the insufficient training data, which is common in medical image analysis, may cause critical gradient vanishing and over-fitting problems to a complicated model with deeper network depth.  Although there are some strategies such as data augmentation, normalization, transfer learning and dropout operation have been proposed to alleviate this problem, the training of the DCNN remains difficult.  

The aforementioned issues motivate us to ease the DCNN training and enhance generalization ability. One promising line of exploration is deeply supervised learning (DSL), which lays emphasis on the intermediate feature representations \cite{ref10,ref11,ref12}. Compared to conventional training scheme that the supervision is only added to the last layer of the network, DSL adds some auxiliary classifiers in the intermediate layers of the network and adopts the joint optimization of auxiliary probabilistic prediction together with the original one. DSL has been verified to be notably effective in combating the gradient vanishing caused by the long-path backpropagation and has also been explored in medical image analysis \cite{ref13,ref14,ref15}. In addition, for many medical specialties, there are insignificant structure and shape differences between classes compared to natural images. Due to the intra-class dissimilarity and inter-class similarity, many studies applied multi-scale feature fusion methods to combine low-level features in the shallower layers and high-level features in the deeper layers \cite{ref16,ref17,ref18}. We deem that via enhancing the gradients flowed back from different auxiliary branches of CNN models, DSL can also realize the fusion of low-level features and high-level features, making it efficient for medical image classification.

However, existing methods for medical image classification simply use the architecture of Deeply-Supervised Nets (DSN) \cite{ref10} with the same or fixed weight for each auxiliary branch, which have not tapped the full potential of DSL to better fuse the multi-level features and comprehensively synergize the knowledges (namely the class probability outputs) extracted from diverse branches. 

In this paper, We revisit the DSL methods and analyze the effectiveness of DSL for medical image classification tasks. For further enhancement of DSN, we propose a deeply supervised layer selective attention network (LSANet), to comprehensively uses label information in feature-level and prediction-level supervision, as illustrated in Figure \ref{fig1}. Inspired by visual attention mechanism \cite{ref19,ref20,ref21}, we propose a novel layer selective attention (LSA) for the first time to adjust the weights of the features in various stages dynamically. Unlike general channel-wise or spatial attention used for feature recalibration in the single layer of CNN models, LSA concentrates on the interaction and selection of the features in different layers of DCNN. For prediction-level supervision, we adopt knowledge synergy strategy using a pairwise probabilistic distribution matching loss \cite{ref11}. This strategy resembles a self-adaptive knowledge incorporation process for optimization consistency. We extensively evaluated our proposed method on the publicly available MedMNIST dataset \cite{ref22}, which contains a variety of medical specialties, and our method has significant improvement of state-of-the-art (SOTA) CNN architectures and visual attention modules. The main contributions of this paper are as follows:

\begin{itemize}
\item We empirically verify that deeply supervised learning is impressively effective for medical image analysis and exhibits terrific performance improvement in classification task. 

\item We pioneer the usage of layer selective attention. LSA dynamically generates different weights for feature maps in different stages of DCNN models, and provides a weight allocation scheme for DSL, so as to facilitate the fusion and selection of low-level and high-level features for medical image classification. 

\item To better leverage label information, we propose a novel LSANet by assembling auxiliary classifiers with LSA and introducing a knowledge synergy strategy, so as to realize feature-level and prediction-level supervision of deep CNN models.

\item Extensive experiments with promising results on public MedMNIST dataset validate the effectiveness of our method for medical image classification. The proposed method provides a creative approach for label-efficient learning in the area of medical image analysis.

\end{itemize}

\section{Related Work}
\noindent\textbf{Deeply Supervised Learning.}
The DSL methodology was established to accelerate convergence and
combat gradient vanishing problem. DSN successfully used auxiliary classifiers in the intermediate layers of the network to train deep CNN models for image classification tasks \cite{ref10}. Similarly, Szegedy et al. \cite{refgooglenet} applied two additional classifiers to optimize the training process of the deep GoogLeNet. DKS \cite{ref11} and DHM \cite{ref12} went one step further by enabling explicit information interactions among all probability outputs of various supervision branches.

In recent years, DSL methods has also been used in the field of medical image analysis. Lei et al. \cite{ref13} utilized deep supervised residual network (DSRN) to recognize HEp-2 cells. The proposed method outperformed the traditional DCNN methods and delivered state-of-the-art performance. Wang et al. \cite{ref23} proposed an innovative 3D convolutional network with densely deep supervision for automated breast ultrasound (ABUS). Extensive experimental results demonstrated their proposed method realized high sensitivity with low false positives. To Accurately count the number of cells in microscopy images, He et al. \cite{ref15} employed auxiliary convolutional neural networks (AuxCNNs) to assist in the training of the designed concatenated fully convolutional regression network (C-FCRN). However, to the best of our knowledge, few studies explored on the proper weight assignment of different auxiliary branches or incorporated the knowledge learned by various classifiers.

\noindent\textbf{Visual Attention Mechanism.}
Attention mechanism is able to focus on meaningful features and inhibiting irrelevant information, thus has been widely applied in the construction of deep CNN models. SENet developed a channel-wise attention mechanism to recalibrates channel-wise dependencies and generate attentive features \cite{ref19}. Woo et al. \cite{ref20} combined channel-wise attention with spatial attention to rescale the importance of different channel and positions.
Base on well-known neuroscience theories, Yang et al. \cite{ref24} proposed a conceptually simple but effective attention module to infer 3-D attention weights for the feature map in a layer of CNN.

Visual attention based methods have also aroused the interest of researchers in the field of medical image analysis. Xing et al. \cite{ref26} proposed a two-branch Attention Guided Deformation Network (AGDN) for wireless capsule endoscopy (WCE) image classification. Cao et al. \cite{ref6} utilized attention feature pyramid network (AttFPN) to realize automatic detection of abnormal cervical cells, which mimicks the way pathologists reading a cervical cytology image. For the classification of multi-modal retinal images, a novel modality-specific attention network (MSAN) was designed to effectively extract the modality-specific diagnostic features from fundus and OCT images \cite{ref27}. However, the existing visual attention based methods aim at improving the joint encoding of spatial and channel information in a single layer of deep CNN models, which haven't considered relationship between the feature maps of different layers. Since multi-scale feature fusion methods have been universally used for medical image analysis, it demonstrates low-level features in shallow layer are also important. Thus, it's a intuitive idea to recalibrate the attention weights of feature maps in different layers.

\section{Methodology}
In this section, we present the formulation of DSL and reveal its potential for medical image classification. Furthermore, we elaborate on our proposed network, illustrate the improved optimization objective and highlight its insight.

\begin{figure*}[t]
\centering
\includegraphics[width=1\textwidth]{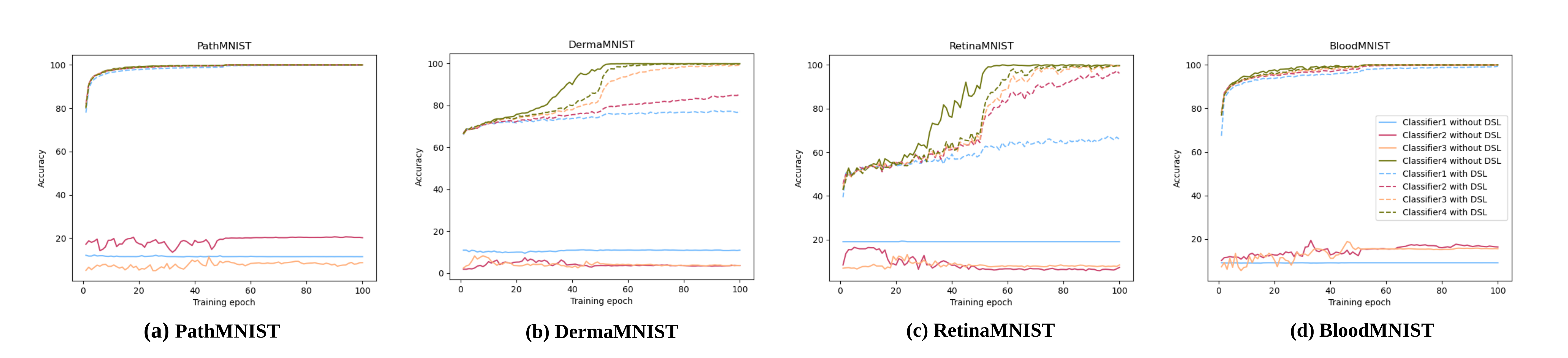} 
\caption{The training results of two sets of classifiers: 4 solid lines present the training accuracy of classifers using the conventional training scheme and 4 dashed lines represent the training accuracy of classifers using DSL.}
\label{fig2}
\end{figure*}

\subsection{Analysis of Deeply Supervised Learning}
\noindent\textbf{Formulation.}
Given an annotated dataset $ \mathcal{S} = \{(x_{i},y_{i}) \mid i = 1, \dots, N\}$ including $N$ training samples collected from $C$ predefined classes where $x_{i}$ denotes the $i^{th}$ sample and $y_{i} \in\{1,2, \ldots, C\}$ is the corresponding ground truth label. Let $f_{M}\left(\boldsymbol{W}_{M}, x_{i}\right)$ be the $C$-dimensional probability output vector for the training sample $x_{i}$ and $\boldsymbol{W}_{M}$ be the learnable weights of an $M$-layer deep CNN model. Then, with supervision signal only attached to the final layer of the network, the optimization objective of the standard training scheme can be defined as
\begin{equation} 
\underset{\boldsymbol{W}_{M}}{\arg\min }  \mathcal{L}_{M}\left(\boldsymbol{W}_{M} ; \mathcal{S}\right) + \lambda \mathcal{R}\left(\boldsymbol{W}_{M}\right),
\label{equation1}
\end{equation}
where $\mathcal{L}_{M}$ is the overall loss of training samples, $\mathcal{R}$ is the regularization term with a weighting factor $\lambda$. Specifically, $\mathcal{L}_{M}$ is
typically defined as the cross-entropy cost function
\begin{equation} 
\mathcal{L}_{M} = - \frac{1}{N} \sum_{i=1}^{N} y_{i} \log f_{M}\left(\boldsymbol{W}_{M}; x_{i}\right).
\label{equation2}
\end{equation}
Most of the current prevailing CNN models adopt this training scheme. Instead, DSN \cite{ref10} provides another training scheme, deeply supervised learning. To create a more transparent learning process, DSN introduces auxiliary classifiers to all hidden layers of the network so that gradients not merely depends on the long-distance backpropagation over stacked modules, but can flow back through its nearest supervision branch as well. Consider $\mathcal{W}_{m}=\{\boldsymbol{W}_{m}^{l} \mid l=1,2, \ldots, L-1\}$ as a set of  weight matrices collected from auxiliary classifiers in hidden layers and $\boldsymbol{W}_{m}^{l}$ as the weight matrix of the $l^{th}$ auxiliary classifier. Since $\lambda \mathcal{R}$ is a default term and has no relation to supervision signals, we omit this term in the following analysis. Then, the optimization objective of the deeply supervised learning scheme can be expressed as
\begin{equation} 
\underset{\boldsymbol{W}_{M},\mathcal{W}_{m}}{\arg\min }  \mathcal{L}_{M}\left(\boldsymbol{W}_{M}; \mathcal{S}\right) + \mathcal{L}_{m}\left(\boldsymbol{W}_{M},\mathcal{W}_{m}; \mathcal{S}\right),
\label{equation3}
\end{equation}
where $\mathcal{L}_{m}$ is the sum of the losses from all auxiliary classifiers with $ \alpha_{l}$ being the weighting factor of the $l^{th}$ auxiliary classifier. Let $f_{m}^{l}(\boldsymbol{W}_{M}, \boldsymbol{W}_{m}^{l}; x_{i})$ be the probabilistic prediction from the $l^{th}$ auxiliary classifier. Then, $\mathcal{L}_{m}$ can be calculated as
\begin{equation} 
\mathcal{L}_{m} = - \frac{1}{N} \sum_{i=1}^{N} \sum_{l=1}^{L-1} \alpha_{l} y_{i} \log f_{m}^{l}(\boldsymbol{W}_{M}, \boldsymbol{W}_{m}^{l}; x_{i}).
\label{equation4}
\end{equation}
By introducing auxiliary loss in Equation \ref{equation4}, the deeply supervised learning scheme allows the network to aggregate gradients from all supervision branches, which is an effective way to combat the gradient vanishing problem.

\noindent\textbf{Potential in medical image classification.}
In order to delve the potential of deep supervised learning (DSL) for medical image classification task, a comparative experiment based on ResNet-50 is performed to explore the interpretability of DSL. Specifically, since ResNet-50 has four stages of stacked residual blocks and the original classifier is only added after the last stage, we attach auxiliary classifiers to the other three stages. All auxiliary classifiers have the same structure consisted of a global average pooling layer and a fully connected layer. We trained two sets of models on four subset of MedMNIST separately. For the model with DSL strategy, the gradients flowed back through four branches simultaneously and the loss was calculated as Equation \ref{equation4}. As for the model without DSL strategy, the loss was only calculated for the classifer in the last stage as Equation \ref{equation2}. We trained the models for 100 epochs  to make them enter into fully saturated zone. Figure \ref{fig2} illustrated the results of the training process.

It can be seen that without DSL, the accuracy of the first three classifiers is quite low and there is hardly effect to discriminate medical images. In addition, the solid blue line, which represents the classifier in the first stage of ResNet-50, is rather smooth with almost no fluctuation, indicating that the gradient has little influence on it after a long-path travel. Using DSL strategy, all classifiers incline to be more discriminative through iterative training. We deem that DSL serves as a proxy to influence the quality of those intermediate layer feature maps, thus to further favor highly discriminative feature maps in final output layer.

In the field of medical image analysis, features in relatively shallow layers are also useful for the final classification. In this paper, we discover that DSL seeks the better output performance while also requiring the satisfactory performance on the part of features in the hidden layers. Thus, DSL is an effective way to integrate multi-level features and improve the performance of medical image classification.

\subsection{Deeply Supervised LSANet}
To fully exploit the deeply supervised learning, we propose a novel deeply supervised layer selective network to perform feature-level and prediction-level supervision. For feature-level supervision, we assign all classifiers a learnable weight value which can be adjusted self-adaptively via LSA. For prediction-level supervision, we use a knowledge matching loss to relieve the optimization inconsistency. Following the definition of Equation \ref{equation3}, the optimization objective of LSANet is defined as
\begin{equation}
\underset{\boldsymbol{W}_{M},\mathcal{W}_{m}}{\arg\min }  \mathcal{L}_{b}\left(\boldsymbol{W}_{M},\mathcal{W}_{m}; \mathcal{S}\right) + \mathcal{L}_{k}\left(\boldsymbol{W}_{M},\mathcal{W}_{m}; \mathcal{S}\right),
\label{equation5}
\end{equation}
where ${L}_{k}$ denotes the loss of prediction-level supervision and  ${L}_{b}$ is the loss of feature-level supervision consisting of ${L}_{M}$ and ${L}_{m}$ with a dynamic weighting factor $\beta$ for each classifier.

\noindent\textbf{Feature Embedding.}
Considering the shape of feature maps in different stages is different, the feature maps in shallow layers have fewer channels and larger resolution. By contrast, the feature maps in deep layers have more channels and smaller resolution. In order to ensure the shape consistency of features at different layers, we use a feature embedding module, as shown in Figure \ref{fig1}. Specifically, for feature maps at different layers, a convolution layer with different kernel size $k$ (the deeper the layer, the smaller the kernel size) is used to harmonize the size of feature maps, and then a subsequent $1\times1$ convolution is used for downsampling to maintain channel consistency for feature maps at different stages. Finally, each stage's feature maps are embed into tensors with the same shape, as the input of layer selective attention.

\noindent\textbf{Layer Selective Attention.}
Since features at different stages have different degrees of influence on the final classification results, when calculating the total loss, the probability outputs of classifiers at these layers cannot be simply given the same weight. In addition, datasets are so diverse and varied in size and specialties that it is difficult to assign a fixed weight to these classifiers for wide application. Furthermore, model training is a dynamic and continuous process. Features in the same depth may have different effects on the final classification results in different training periods. Thus, the core contribution of layer selective attention is introducing a weight allocation method for deeply supervised learning and ensuring its generalization.

After feature embedding, we acquire four input feature maps and each denotes as $\mathbf U  \in \mathbb{R}^{H\times W \times C}$. We first concatenate them and aggregate the spatial information for each channel as $\mathbf U_{avg} \in \mathbb{R}^{1\times 1 \times 4C}$ by using average pooling. Next, the condensed tensor is weighted by $MLP$ consisted of two cascaded fully connected layer with a ReLU activation bewteen them. The whole operation process is computed as
\begin{small}
\begin{equation} 
 \mathbf G(\mathbf U) = \sigma(MLP(Avgpool(\mathbf U))) = \sigma(\mathbf W_{2}\delta(\mathbf W_{1}\mathbf U_{avg})),
\label{equation6}
\end{equation}
\end{small}
where $\sigma$ denotes a sigmoid function, $\delta$ is a ReLU function, $\mathbf W_{1}\in\mathbb{R}^{\frac{C}{r}\times C}$ and $\mathbf W_{2}\in\mathbb{R}^{ C \times \frac{C}{r}}$. We use a reduction ratio $r$ to reduce the parameters. After the channel attention is calculated, we split the output tensor into four tensors, each as  $\mathbf z  \in \mathbb{R}^{1\times 1 \times C}$. Then, we calculate each output tensor's mean and finally use a softmax function to acquire the layer selective attention weight $\beta$. After the LSA assigns weights, the loss of feature-level supervision ${L}_{b}$ is defined as follows
\begin{equation} 
\mathcal{L}_{b} = - \frac{1}{N} \sum_{i=1}^{N} \sum_{l=1}^{L} \beta_{l} y_{i} \log f_{b}^{l}(\boldsymbol{W}_{M}, \boldsymbol{W}_{m}^{l}; x_{i}),
\label{equation7}
\end{equation}
where $f_{b}^{l}(\boldsymbol{W}_{M}, \boldsymbol {W}_{m}^{l}; x_{i})$ represents the probabilistic protection of each classifier. $\beta_{l}$ not only weights the auxiliary classifiers but also weights the raw classifier on top of the last layer, and can be updated dynamically during the whole training process. LSA restricts attention to the parts of feature space, lead to highly discriminative hidden layer feature maps and further lead to higher performance of final output.

\noindent\textbf{Knowledge Synergy Strategy}
Knowledge Synergy strategy can facilitate aggregation of hierarchical knowledge (extracted from different classifiers) and advance information consistency among them. Considering there are $L$ classifiers where the first $L-1$ classifiers represent additional classifiers in auxiliary branches and the last one is the original classifier on top of the last layer. The knowledge matching between any two classifiers is a modified KL divergence

\begin{equation}
\mathcal{L}_{k}=-\frac{1}{N} \sum_{i=1}^{N} \sum_{p=1}^{L} \sum_{\substack{q=1 \\ q \neq p}}^{L}\mu_{p q} f_{p} \log \frac{f_{p}}{f_{q}},
\label{equation8}
\end{equation}
where $f_{p}$ and $f_{q}$ are the class probability outputs of the classifier $p$ and $q$, and $\mu_{p q}$ weights the loss of the pairwise knowledge matching from the classifier $p$ to $q$, which is set as 1 for all pairs of knowledge matching in this study. In this way, classifier $p$ inclines to mimic the classifier $q$ and the knowledge can be spread in all classifiers.

\section{Experiments}
\subsection{Datasets and Experiment Settings}
\noindent\textbf{Dataset.}
In this study, we perform all experiments on a public medical classification dataset, MedMNIST \cite{ref7}. MedMNIST is a large-scale benchmark dataset for biomedical image analysis and covers a large number of medical specialties. We evaluate our proposed LSANet with four color image subsets of MedMNIST, i.e., PathMNIST, DermaMNIST, RetinaMNIST and BloodMNIST.
\begin{itemize}
\item \textbf{PathMNIST} provites 100,000 training images and 7180 test images comprised of 9 types of tissues for predicting survival from colorectal cancer.
\item \textbf{DermaMNIST} consists of 10,015 dermatoscopic images categorized as 7 different pigmented skin lesions.
\item \textbf{RetinaMNIST} contains 1,600 retina fundus images used for 5-level grading of diabetic retinopathy severity.
\item \textbf{BloodMNIST} collects a total of 17,092 images and is organized into 8 classes for the classification of individual normal cells.
\end{itemize}
\noindent\textbf{Experiment Settings.}
Our proposed network is implemented with PyTorch and all experiments are conducted on a work station with 4 RTX3090 (about 24GB) GPUs. We use the resized input image resolution (224 $\times$ 224), batch size (64) , training epochs (100) and optimizer (Adam with an initial learning rate of 0.001) for training. We choose the accuracy (ACC) and the area under the curve (AUC) of receiver operating characteristic (ROC) as the metrics to evaluate our model for all experiments. For each network, we run each method 3 times and report mean ACC and AUC.

\noindent\textbf{Implementation Details.}
In this work, we use VGGNet \cite{refvgg}, ResNet \cite{refresnet}, Wide Residual Networks (WRN) \cite{ref30}, ResNeXt \cite{ref31} and DenseNet \cite{ref32} as the backbones to build LSANet and perform contrastive experiments. For VGGNet, we attach classifiers (including three additional classifiers and an original classifier) on top of last four max pooling layers. For ResNet, WRN and ResNeXt, we add three auxiliary classifiers on top of the block Conv2\_x, Conv3\_x, Conv4\_x. As for DenseNet, we remain the raw classifier and append three auxiliary classifiers on top of the first dense block, second dense block and third dense block. While constructing DSN and LSANet, we add three auxiliary classifiers and all classfiers consist of a global average pooling layer and a fully connected layer. In regard to DKS, there are only two auxiliary classifiers on the top of Conv3\_x and Conv4\_x for ResNet.

\begin{table*}[t]
\caption{Comparison of different deeply supervised learning method on MedMNIST dataset.}
\small
\centering
\begin{tabular}{llllllllll}
\toprule
\multirow{2}{*}{Model}      & \multirow{2}{*}{Method} & \multicolumn{2}{l}{PathMNIST} & \multicolumn{2}{l}{DermaMNIST} & \multicolumn{2}{l}{RetinaMNIST} & \multicolumn{2}{l}{BloodMNIST} \\
\cline{3-10}
                            &                         & ACC           & AUC           & ACC            & AUC           & ACC            & AUC            & ACC            & AUC           \\
\midrule
\multirow{4}{*}{ResNet-18}  & Baseline                & 89.87         & 98.90         & 74.56          & 92.05         & 51.50          & 71.55          & 96.14          & 99.79         \\
                            & DSN                     & 91.36         & 98.97         & 76.66          & 92.20         & 54.25          & 72.98          & 97.11          & 99.85         \\
                            & DKS                     & 92.40         & \textbf{99.24}         & 75.96          & 92.23         & 53.50          & 72.41          & 96.49          & 99.82         \\
                            & LSANet                  & 91.88         & 99.11         & 78.10          & 93.69         & 56.00          & 72.73          & 97.34          & \textbf{99.88}         \\
\midrule
\multirow{4}{*}{ResNet-34}  & Baseline                & 90.33         & 99.12         & 74.01          & 91.39         & 52.25          & 70.77          & 96.02          & 99.78         \\
                            & DSN                     & 91.23         & 99.15         & 75.71          & 92.06         & 54.25          & 71.57          & 97.05          & 99.87         \\
                            & DKS                     & 91.89         & 99.23         & 75.41          & 91.93         & 53.75          & 73.68          & 96.35          & 99.82         \\
                            & LSANet                  & 91.75         & 99.23         & \textbf{78.70}          & \textbf{93.96}         & \textbf{56.75}          & 73.46          & \textbf{97.40}          & \textbf{99.88}         \\
\midrule
\multirow{4}{*}{ResNet-50}  & Baseline                & 90.29         & 98.54         & 73.67          & 91.13         & 52.00          & 71.21          & 95.56          & 99.76         \\
                            & DSN                     & 92.13         & 99.19         & 75.11          & 91.28         & 53.50          & 72.62          & 96.78          & 99.84         \\
                            & DKS                     & 92.68         & 99.07         & 75.21          & 91.29             & 53.25          & 72.95          & 96.08          & 99.81         \\
                            & LSANet                  & \textbf{92.74}         & 99.12         & 78.15          & 93.08         & 54.50          & \textbf{73.84}          & 97.11          & 99.86         \\
\midrule
\multirow{4}{*}{ResNet-101} & Baseline                & 91.27         & 98.78         & 73.27          & 90.92         & 50.50          & 70.59          & 95.67          & 99.74         \\
                            & DSN                     & 92.19         & 99.12         & 75.06          & 91.37         & 53.00          & 71.67          & 96.87          & 99.85         \\
                            & DKS                     & 92.47         & 99.03         & 74.86          & 91.66         & 52.75          & 71.16          & 95.94          & 99.78         \\
                            & LSANet                  & 92.70         & 99.15         & 78.35          & 93.66         & 54.25          & 72.96          & 97.22          & 99.86         \\
\bottomrule
\end{tabular}
\label{table1}
\end{table*}

\subsection{Main Result}
We compared LSANet with other two deeply supervised learning methods, DSN \cite{ref10} and DKS \cite{ref11} on MedMNIST datasets using the backbone of ResNet. The comparative results are shown in Table \ref{table1}. 

Compared with baseline (original ResNet), the performances of all three DSL methods are significantly improved, indicating that DSL is very effective for medical image classification tasks. In particular, on DermaMNIST, RetinaMNIST and BloodMNIST datasets, LSANet performs much better than DSN and DKS in all four constructions of ResNet with different depths. However, we also discover that on PathMNIST dataset, the performance of DKS is relatively prominent, which may be caused by the following reasons: (1) PathMNIST has much more images than the other three datasets, which makes the number of parameters of ResNet with fewer layers (less than 34 layers) not enough to fit the data distribution. (2) DKS adds the same residual block as the main branch to the auxiliary branch while building the network, thus introduces a lot of extra helpful parameters to fit the large dataset, PathMNIST. As the network deepens, such as in ResNet-50, the performance of LSANet becomes comparable to DKS mainly because When the number of parameters is sufficient or more, LSANet is more effective in solving the problem of gradient disappearance and parameter redundancy, and can make better use of the low-level features and high-level features in different depths of CNN models for medical image analysis. Generally speaking, large datasets can be fitted by building deeper networks. When the datasets are relatively small, DSL is a very helpful approach to make more comprehensive use of label information to achieve label-efficient learning for medical image classification.

\subsection{Ablation Studies}
\noindent\textbf{Influences of different implementations of LSA.} 
In order to better obtain the attention weight of feature maps in different stages, we conduct an ablation study on DermaMNIST to investigate different implementations of layer selective attention. Specifically, based on ResNet-34, we inspect the influence of two cascaded fully connected layer with different reduction ratio \cite{ref19}, 1D convolution layer with different kernel size \cite{ref21}, gated attention \cite{ref29} and self attention \cite{ref28} to realize LSA. Table \ref{table2} exhibits the experimental results. Two fully connected layer with a reduction ratio of eight posseses the best performance while self attention method seems inappropriate for the construction of LSA.
\begin{table}[t]
\caption{Ablation studies on different implementation of layer selective attention.}
\small
\centering
\begin{tabular}{llll}
\toprule
Attention   Type                        & Operation            & ACC            & AUC            \\
\midrule
\multirow{3}{*}{Two FC Layer} & reduction ratio = 4  & 78.00          & 93.72          \\
                                        & reduction ratio = 8  & \textbf{78.70} & \textbf{93.96} \\
                                        & reduction ratio = 16 & 78.55          & 93.93          \\
\midrule
\multirow{3}{*}{1D Convolution}         & kernel size = 3      & 78.35          & 94.01          \\
                                        & kernel size = 5      & 78.25          & 93.53          \\
                                        & kernel size = 7      & 78.40          & 93.66          \\
\midrule
Gated Attention                         & -                    & 78.40          & 93.85          \\
Self Attention                          & -                    & 77.81          & 93.64          \\
\bottomrule
\end{tabular}
\label{table2}
\end{table}

\noindent\textbf{Analysis of two-level supervision.}
To demonstrate the effectiveness of each proposed component, on DermaMNIST dataset, we conduct experiments by using ResNet-34 as the backbone to examine the feature-level supervision (realized by LSA) and prediction-level supervision (realized by KS). According to the results shown in Table \ref{table3}, we can see that feature-level supervision and prediction-level supervision obtain 2.70\% and 3.45\% increase in classification accuracy respectively, when compared with the basic DSN. Furthermore, LSANet, which realizes more comprehensive supervision by the joint use of DSN and LSA, presents the highest performance.
\begin{table}[t]
\caption{Ablation studies on Feature-level and Prediction-level supervision, using different optimization objective.}
\small
\centering
\begin{tabular}{lllllll}
\toprule
Method   & DSL & LSA & KS & ACC            & AUC      &Gain      \\
\midrule
Baseline &    &    &   & 74.01          & 91.39       &-   \\
DSN      & \Checkmark   &    &   & 75.71          & 92.06     &1.70/0.67     \\
DSN+KS   & \Checkmark   &    & \Checkmark  & 77.46          & 93.80     &3.45/2.41     \\
DSN+LSA  & \Checkmark   & \Checkmark   &   & 76.71          & 92.48     &2.70/1.09     \\
LSANet   & \Checkmark   & \Checkmark   & \Checkmark  &\textbf{78.70} &\textbf{93.96}  &\textbf{4.69/2.57} \\
\bottomrule
\end{tabular}
\label{table3}
\end{table}

\noindent\textbf{Comparison of different combination of output nodes.}
Given a backbone network, the configuration scheme of auxiliary classifiers are crucial for the deeply supervised learning methods. We consider different settings by adding classifiers to at most three intermediate layer locations (including the block Conv2\_x, Conv3\_x and Conv4\_x) of ResNet-34. Detailed results are shown in Table \ref{table4} where $C_{1}$, $C_{2}$, $C_{3}$ and $C_{4}$ denote the output nodes (classifiers) which are connected on top of the block Conv2\_x, Conv3\_x, Conv4\_x and the last layer, sequentially. From Table 4, we observe that the more auxiliary classifiers, the better the accuracy improvement. Due to the use of layer selective attention, all output nodes can adjust the weight adaptively to seek more suitable feature space while inhibiting bad feature representations. Therefore, more auxiliary classifiers provide bigger feature searching space and present higher performance.
\begin{table}[t]
\caption{Accuracy and AUC gains of LSANet with auxiliary classifiers connected to different intermediate layers of ResNet-34.}
\small
\centering
\begin{tabular}{lllll}
\toprule
Model    & ACC            & AUC            & Gain      \\
\midrule
Baseline( $C_{4}$) & 74.01          & 91.39      &-    \\
$C_{1}C_{4}$       & 75.46          & 92.35       &1.45/0.96   \\
$C_{2}C_{4}$       & 76.41          & 93.22       &2.40/1.83   \\
$C_{3}C_{4}$       & 76.16          & 91.98       &2.15/0.59   \\
$C_{1}C_{2}C_{4}$      & 76.46          & 92.78    &2.45/1.39      \\
$C_{1}C_{3}C_{4}$      & 77.41          & 93.41    &3.40/2.02      \\
$C_{2}C_{3}C_{4}$      & 77.46          & 93.22    &3.45/1.83     \\
$C_{1}C_{2}C_{3}C_{4}$     & \textbf{78.70} & \textbf{93.96}  &\textbf{4.69/2.57} \\
\bottomrule
\end{tabular}
\label{table4}
\end{table}

\noindent\textbf{Investigation of multiple mainstream CNN models.}
We validate the universality and generalizability of LSANet on a variety of mainstream CNN models including VGGNet \cite{refvgg}, Wide Residual Networks (WRN) \cite{ref30}, ResNeXt \cite{ref31} and DenseNet \cite{ref32}. The detailed results on DermaMNIST dataset are shown in Table \ref{table5}. With regard to different CNN architectures, LSANet can always be superior to its corresponding counterparts, no matter it is VGGNet without residual connection nor DenseNet with sophisticated connection path. Particularly, DenseNet-169 has the best recognition performance with an accuracy of 79.05\% and AUC of 93.73\%  after the use of LSANet. These experiments clearly validate the effectiveness of the proposed method when training prevailing CNN models. 

\begin{table}[t]
\caption{Results of LSANet when applied to multiple CNN architectures.}
\scriptsize
\centering
\begin{tabular}{llll}
\toprule
Model       & Method   & ACC            & AUC            \\
\midrule
VGG-16       & Baseline & 72.87          & 90.99          \\
            & LSANet      & \textbf{76.16} & \textbf{92.92} \\
VGG-19       & Baseline & 72.27          & 89.81          \\
            & LSANet      & \textbf{76.51} & \textbf{92.51} \\
\midrule
WRN-50       & Baseline & 74.06          & 90.58          \\
            & LSANet     & \textbf{77.11} & \textbf{92.61} \\
WRN-101      & Baseline & 73.97          & 90.54          \\
            & LSANet      & \textbf{76.51} & \textbf{93.00} \\
\midrule
ResNeXt-50   & Baseline & 74.61          & 91.66          \\
            & LSANet      & \textbf{78.60} & \textbf{93.66} \\
ResNeXt-101  & Baseline & 74.31          & 91.27          \\
            & LSANet      & \textbf{78.45} & \textbf{93.72} \\
\midrule
DenseNet-121 & Baseline & 76.11          & 91.27          \\
            & LSANet      & \textbf{79.05} & \textbf{93.62} \\
DenseNet-169 & Baseline & 76.26          & 92.20          \\
            & LSANet      & \textbf{78.95} & \textbf{93.73} \\
DenseNet-201 & Baseline & 75.76          & 91.54          \\
            & LSANet      & \textbf{78.85} & \textbf{94.03} \\
\bottomrule
\end{tabular}
\label{table5}
\end{table}

\noindent\textbf{Investigation of various visual attention modules.}
We further conduct an ablation study to investigate the combination of LSANet and various visual attention modules on DermaMNIST dataset. To be specific, we use ResNet-34 as the backbone, and plug SE module \cite{ref19}, CBAM \cite{ref20}, ECA \cite{ref21}, SRM \cite{ref33} and SimAM \cite{ref24} into it. The results are shown in \ref{table6}. It can be observed from Table \ref{table6} that the employment of visual attention modules greatly improve the accuracy of the original ResNet-34. In addition, ResNet-34 with SimAM achieves the state-of-the-art performance over all other models by the conbination of LSANet. The experimental results demonstrate that LSANet can be effectively used with the attention module to make the model achieve better performance for medical image classification.

\begin{table}[t]
\caption{Results of LSANet when applied to different visual attention modules using DermaMNIST dataset.}
\small
\centering
\begin{tabular}{llll}
\toprule
Attention Module & Method   & ACC            & AUC            \\
\midrule
Original         & Baseline & 74.01          & 91.39          \\
                 & LSANet   & \textbf{78.70} & \textbf{93.96} \\
+SE              & Baseline & 75.06          & 92.25          \\
                 & LSANet   & \textbf{78.80} & \textbf{94.06} \\
+CBAM            & Baseline & 74.86          & 92.29          \\
                 & LSANet   & \textbf{78.90} & \textbf{93.99} \\
+ECA             & Baseline & 75.41          & 92.61          \\
                 & LSANet   & \textbf{79.15} & \textbf{94.19} \\
+SimAM           & Baseline & 74.81          & 91.62          \\
                 & LSANet   & \textbf{79.40} & \textbf{94.42} \\
+SRM             & Baseline & 76.16          & 92.84          \\
                 & LSANet   & \textbf{79.20} & \textbf{94.39} \\
\bottomrule
\end{tabular}
\label{table6}
\end{table}

\section{Conclusion}
In this paper, we we analyze the rationality and effectiveness of deeply supervised learning to alleviate gradient vanishing and parameter redundancy problems, and tap the excellent potential of deeply supervised learning in medical image classification. We demonstrate that deeply supervised learning can realize the interactive fusion of low-level and high-level features. Through delving into the dynamic training schemes of deep supervision, a novel network called LSANet is proposed to comprehensively utilizes label information in multi-level supervision for small medical datasets so as to realize label-efficient learning for medical image analysis. Extensive experiments on public Medmnist dataset validate that this approach is beneficial to improving the accuracy and generalization ability of deep neural networks on various medical image classification tasks.

\bibliographystyle{IEEEbib}
\bibliography{Ref_LSANet}

\begin{thebibliography}{10}

\bibitem{ref1}
Andre Esteva, Brett Kuprel, Roberto~A Novoa, Justin Ko, Susan~M Swetter,
  Helen~M Blau, and Sebastian Thrun,
\newblock ``Dermatologist-level classification of skin cancer with deep neural
  networks,''
\newblock {\em nature}, vol. 542, no. 7639, pp. 115--118, 2017.

\bibitem{ref2}
Jeffrey De~Fauw, Joseph~R Ledsam, Bernardino Romera-Paredes, Stanislav Nikolov,
  Nenad Tomasev, Sam Blackwell, Harry Askham, Xavier Glorot, Brendan
  O’Donoghue, Daniel Visentin, et~al.,
\newblock ``Clinically applicable deep learning for diagnosis and referral in
  retinal disease,''
\newblock {\em Nature medicine}, vol. 24, no. 9, pp. 1342--1350, 2018.

\bibitem{ref3}
Jeroen Van~der Laak, Geert Litjens, and Francesco Ciompi,
\newblock ``Deep learning in histopathology: the path to the clinic,''
\newblock {\em Nature medicine}, vol. 27, no. 5, pp. 775--784, 2021.

\bibitem{ref4}
Andre Esteva, Alexandre Robicquet, Bharath Ramsundar, Volodymyr Kuleshov, Mark
  DePristo, Katherine Chou, Claire Cui, Greg Corrado, Sebastian Thrun, and Jeff
  Dean,
\newblock ``A guide to deep learning in healthcare,''
\newblock {\em Nature medicine}, vol. 25, no. 1, pp. 24--29, 2019.

\bibitem{refvgg}
Karen Simonyan and Andrew Zisserman,
\newblock ``Very deep convolutional networks for large-scale image
  recognition,''
\newblock {\em arXiv preprint arXiv:1409.1556}, 2014.

\bibitem{refgooglenet}
Christian Szegedy, Wei Liu, Yangqing Jia, Pierre Sermanet, Scott Reed, Dragomir
  Anguelov, Dumitru Erhan, Vincent Vanhoucke, and Andrew Rabinovich,
\newblock ``Going deeper with convolutions,''
\newblock in {\em Proceedings of the IEEE conference on computer vision and
  pattern recognition}, 2015, pp. 1--9.

\bibitem{refresnet}
Kaiming He, Xiangyu Zhang, Shaoqing Ren, and Jian Sun,
\newblock ``Deep residual learning for image recognition,''
\newblock in {\em Proceedings of the IEEE conference on computer vision and
  pattern recognition}, 2016, pp. 770--778.

\bibitem{ref5}
Lequan Yu, Hao Chen, Qi~Dou, Jing Qin, and Pheng-Ann Heng,
\newblock ``Automated melanoma recognition in dermoscopy images via very deep
  residual networks,''
\newblock {\em IEEE transactions on medical imaging}, vol. 36, no. 4, pp.
  994--1004, 2017.

\bibitem{ref6}
Lei Cao, Jinying Yang, Zhiwei Rong, Lulu Li, Bairong Xia, Chong You, Ge~Lou,
  Lei Jiang, Chun Du, Hongxue Meng, et~al.,
\newblock ``A novel attention-guided convolutional network for the detection of
  abnormal cervical cells in cervical cancer screening,''
\newblock {\em Medical image analysis}, vol. 73, pp. 102197, 2021.

\bibitem{ref7}
Jiancheng Yang, Rui Shi, and Bingbing Ni,
\newblock ``Medmnist classification decathlon: A lightweight automl benchmark
  for medical image analysis,''
\newblock in {\em 2021 IEEE 18th International Symposium on Biomedical Imaging
  (ISBI)}. IEEE, 2021, pp. 191--195.

\bibitem{ref8}
Shervin Minaee, Rahele Kafieh, Milan Sonka, Shakib Yazdani, and
  Ghazaleh~Jamalipour Soufi,
\newblock ``Deep-covid: Predicting covid-19 from chest x-ray images using deep
  transfer learning,''
\newblock {\em Medical image analysis}, vol. 65, pp. 101794, 2020.

\bibitem{ref9}
Peng Yao, Shuwei Shen, Mengjuan Xu, Peng Liu, Fan Zhang, Jinyu Xing, Pengfei
  Shao, Benjamin Kaffenberger, and Ronald~X Xu,
\newblock ``Single model deep learning on imbalanced small datasets for skin
  lesion classification,''
\newblock {\em IEEE transactions on medical imaging}, vol. 41, no. 5, pp.
  1242--1254, 2022.

\bibitem{ref10}
Chen-Yu Lee, Saining Xie, Patrick Gallagher, Zhengyou Zhang, and Zhuowen Tu,
\newblock ``Deeply-supervised nets,''
\newblock in {\em Artificial intelligence and statistics}. PMLR, 2015, pp.
  562--570.

\bibitem{ref11}
Dawei Sun, Anbang Yao, Aojun Zhou, and Hao Zhao,
\newblock ``Deeply-supervised knowledge synergy,''
\newblock in {\em Proceedings of the IEEE/CVF Conference on Computer Vision and
  Pattern Recognition}, 2019, pp. 6997--7006.

\bibitem{ref12}
Duo Li and Qifeng Chen,
\newblock ``Dynamic hierarchical mimicking towards consistent optimization
  objectives,''
\newblock in {\em Proceedings of the IEEE/CVF Conference on Computer Vision and
  Pattern Recognition}, 2020, pp. 7642--7651.

\bibitem{ref13}
Haijun Lei, Tao Han, Feng Zhou, Zhen Yu, Jing Qin, Ahmed Elazab, and Baiying
  Lei,
\newblock ``A deeply supervised residual network for hep-2 cell classification
  via cross-modal transfer learning,''
\newblock {\em Pattern Recognition}, vol. 79, pp. 290--302, 2018.

\bibitem{ref14}
Pengfei Gu, Hao Zheng, Yizhe Zhang, Chaoli Wang, and Danny~Z Chen,
\newblock ``kcbac-net: Deeply supervised complete bipartite networks with
  asymmetric convolutions for medical image segmentation,''
\newblock in {\em International Conference on Medical Image Computing and
  Computer-Assisted Intervention}. Springer, 2021, pp. 337--347.

\bibitem{ref15}
Shenghua He, Kyaw~Thu Minn, Lilianna Solnica-Krezel, Mark~A Anastasio, and Hua
  Li,
\newblock ``Deeply-supervised density regression for automatic cell counting in
  microscopy images,''
\newblock {\em Medical Image Analysis}, vol. 68, pp. 101892, 2021.

\bibitem{ref16}
Zongwei Zhou, Md~Mahfuzur~Rahman Siddiquee, Nima Tajbakhsh, and Jianming Liang,
\newblock ``Unet++: Redesigning skip connections to exploit multiscale features
  in image segmentation,''
\newblock {\em IEEE transactions on medical imaging}, vol. 39, no. 6, pp.
  1856--1867, 2019.

\bibitem{ref17}
Imran Razzak and Saeeda Naz,
\newblock ``Unit-vise: deep shallow unit-vise residual neural networks with
  transition layer for expert level skin cancer classification,''
\newblock {\em IEEE/ACM Transactions on Computational Biology and
  Bioinformatics}, 2020.

\bibitem{ref18}
Xiaokun Liang, Na~Li, Zhicheng Zhang, Jing Xiong, Shoujun Zhou, and Yaoqin Xie,
\newblock ``Incorporating the hybrid deformable model for improving the
  performance of abdominal ct segmentation via multi-scale feature fusion
  network,''
\newblock {\em Medical Image Analysis}, vol. 73, pp. 102156, 2021.

\bibitem{ref19}
Jie Hu, Li~Shen, and Gang Sun,
\newblock ``Squeeze-and-excitation networks,''
\newblock in {\em Proceedings of the IEEE conference on computer vision and
  pattern recognition}, 2018, pp. 7132--7141.

\bibitem{ref20}
Sanghyun Woo, Jongchan Park, Joon-Young Lee, and In~So Kweon,
\newblock ``Cbam: Convolutional block attention module,''
\newblock in {\em Proceedings of the European conference on computer vision
  (ECCV)}, 2018, pp. 3--19.

\bibitem{ref21}
Wang Qilong, Wu~Banggu, Zhu Pengfei, Li~Peihua, Zuo Wangmeng, and Hu~Qinghua,
\newblock ``Eca-net: Efficient channel attention for deep convolutional neural
  networks,''
\newblock in {\em Proceedings of the IEEE conference on computer vision and
  pattern recognition}, 2020, pp. 11531--11539.

\bibitem{ref22}
Yang Jiancheng, Shi Rui, Wei Donglai, Liu Zequan, Zhao Lin, Ke~Bilian, Pfister
  Hanspeter, and Ni~Bingbing,
\newblock ``Medmnist v2: {A} large-scale lightweight benchmark for 2d and 3d
  biomedical image classification,'' 2021.

\bibitem{ref23}
Yi~Wang, Na~Wang, Min Xu, Junxiong Yu, Chenchen Qin, Xiao Luo, Xin Yang, Tianfu
  Wang, Anhua Li, and Dong Ni,
\newblock ``Deeply-supervised networks with threshold loss for cancer detection
  in automated breast ultrasound,''
\newblock {\em IEEE transactions on medical imaging}, vol. 39, no. 4, pp.
  866--876, 2019.

\bibitem{ref24}
Lingxiao Yang, Ru-Yuan Zhang, Lida Li, and Xiaohua Xie,
\newblock ``Simam: A simple, parameter-free attention module for convolutional
  neural networks,''
\newblock in {\em International conference on machine learning}. PMLR, 2021,
  pp. 11863--11874.

\bibitem{ref26}
Xiaohan Xing, Yixuan Yuan, and Max Q-H Meng,
\newblock ``Zoom in lesions for better diagnosis: Attention guided deformation
  network for wce image classification,''
\newblock {\em IEEE Transactions on Medical Imaging}, vol. 39, no. 12, pp.
  4047--4059, 2020.

\bibitem{ref27}
Xingxin He, Ying Deng, Leyuan Fang, and Qinghua Peng,
\newblock ``Multi-modal retinal image classification with modality-specific
  attention network,''
\newblock {\em IEEE Transactions on Medical Imaging}, vol. 40, no. 6, pp.
  1591--1602, 2021.

\bibitem{ref30}
Sergey Zagoruyko and Nikos Komodakis,
\newblock ``Wide residual networks,''
\newblock in {\em Proceedings of the British Machine Vision Conference}. 2016,
  {BMVA} Press.

\bibitem{ref31}
Saining Xie, Ross Girshick, Piotr Doll{\'a}r, Zhuowen Tu, and Kaiming He,
\newblock ``Aggregated residual transformations for deep neural networks,''
\newblock in {\em Proceedings of the IEEE conference on computer vision and
  pattern recognition}, 2017, pp. 1492--1500.

\bibitem{ref32}
Gao Huang, Zhuang Liu, Laurens Van Der~Maaten, and Kilian~Q Weinberger,
\newblock ``Densely connected convolutional networks,''
\newblock in {\em Proceedings of the IEEE conference on computer vision and
  pattern recognition}, 2017, pp. 4700--4708.

\bibitem{ref29}
Maximilian Ilse, Jakub Tomczak, and Max Welling,
\newblock ``Attention-based deep multiple instance learning,''
\newblock in {\em International conference on machine learning}. PMLR, 2018,
  pp. 2127--2136.

\bibitem{ref28}
Ashish Vaswani, Noam Shazeer, Niki Parmar, Jakob Uszkoreit, Llion Jones,
  Aidan~N Gomez, {\L}ukasz Kaiser, and Illia Polosukhin,
\newblock ``Attention is all you need,''
\newblock {\em Advances in neural information processing systems}, vol. 30,
  2017.

\bibitem{ref33}
HyunJae Lee, Hyo-Eun Kim, and Hyeonseob Nam,
\newblock ``Srm: A style-based recalibration module for convolutional neural
  networks,''
\newblock in {\em Proceedings of the IEEE/CVF International Conference on
  Computer Vision}, 2019, pp. 1854--1862.

\end{thebibliography}

\end{document}